\pdfoutput=1

\PassOptionsToPackage{dvipsnames}{xcolor}
\documentclass[11pt]{article}
\usepackage[final]{acl}

\usepackage{times}
\usepackage{latexsym}
\usepackage{zshi}
\usepackage{array}
\usepackage{rotating} 

\usepackage[T1]{fontenc}

\usepackage[utf8]{inputenc}

\usepackage{microtype}

\usepackage{inconsolata}

\newcommand{\reducedstrut}{\vrule width 0pt height .9\ht\strutbox depth .9\dp\strutbox\relax}
\newcommand{\myred}[1]{%
  \begingroup
  \setlength{\fboxsep}{0pt}%
  \colorbox{BurntOrange}{\reducedstrut#1\/}%
  \endgroup
}

\newcommand{\mygreen}[1]{%
  \begingroup
  \setlength{\fboxsep}{0pt}%
  \colorbox{LimeGreen}{\reducedstrut#1\/}%
  \endgroup
}

\title{Improving the Generation Quality of Watermarked Large Language Models via Word Importance Scoring}

\author{Yuhang Li\thanks{\hspace{5pt}Alphabetical order}\\
  UMich\\
  \texttt{liyuhang@umich.edu} \\\And
  Yihan Wang$^*$ \\
  UCLA \\
  \texttt{wangyihan617@gmail.com} \\\AND
  Zhouxing Shi \\
  UCLA \\
  \texttt{zshi@cs.ucla.edu} \\\And
  Cho-Jui Hsieh\\
  UCLA\\
  \texttt{chohsieh@cs.ucla.edu}
  }

\begin{document}
\maketitle
\begin{abstract}
The strong general capabilities of Large Language Models (LLMs) bring potential ethical risks if they are unrestrictedly accessible to malicious users. Token-level watermarking inserts watermarks in the generated texts by altering the token probability distributions with a private random number generator seeded by its prefix tokens. However, this watermarking algorithm alters the logits during generation, which can lead to a downgraded text quality if it chooses to promote tokens that are less relevant given the input. In this work, we propose to improve the quality of texts generated by a watermarked language model by Watermarking with Importance Scoring (WIS). At each generation step, we estimate the importance of the token to generate, and prevent it from being impacted by watermarking if it is important for the semantic correctness of the output. We further propose three methods to predict importance scoring, including a perturbation-based method and two model-based methods. Empirical experiments show that our method can generate texts with better quality with comparable level of detection rate. 
\end{abstract}

\section{Introduction}

Recent progress in Large Language Models (LLMs) has demonstrated their strong capabilities of following human instructions and performing general tasks with prompts \citep{openai_chatgpt,chowdhery2022palm,touvron2023llama}. While the strong general capability is useful for many real-world applications, it also leads to potential ethical risks if it is unrestrictedly accessible to malicious usage, such as online spamming, misinformation, or academic plagiarism. Several methods have been proposed to detect texts generated by certain LLMs \cite{mitchell2023detectgpt, solaiman2019release, AITextClassifier, kirchenbauer2023watermark, hou2023semstamp}, among which, token-level watermarking \cite{kirchenbauer2023watermark} can provide fast and relatively robust detection~\cite{shi2023red}.

However, watermarking intervenes the generation process with altered logits values, which can lead to a trade-off between text quality and detection success rate. To improve text generation quality, \citet{kirchenbauer2023watermark} tends to only impact tokens with high logits entropies, with the intuition that tokens with low logits entropies have few alternatives and watermarking these tokens may have a more significant impact on text quality. However, due to the well-known hallucinations in LLMs~\citep{zhang2023siren,yao2023llm,rawte2023survey}, the entropy of logits may not correctly reflect the factual importance of tokens in the generated output.

In this paper, we aim to improve the trade-off between text quality and detection rate of token-level watermarking on LLMs, with a separate module to estimate the importance of tokens in generation. We propose to prioritize tokens with higher estimated importance scores. We further propose two types of inference-time importance scoring functions: A perturbation-base scoring method, which estimates the importance score by comparing the semantic changes after removing the token, and model-based scoring methods, which fine-tune a language model (LM) to predict the importance score with a dataset labeled by counting the appearances of each token in multiple paraphrases.
We show an example with improved generation quality from WIS in Table \ref{tab:example}. We summarize our contributions below:
\begin{itemize}
    \item We evaluate the trade-off between watermark detection and text quality with several text generation datasets.
    \item We propose a framework with a token importance scoring module to improve the trade-off of LLM watermarking, and we propose three importance scoring methods, including a perturbation-based method and two model-based methods.
    \item We demonstrate that the proposed framework can improve the text quality with similar level of watermark detection rate.
\end{itemize}

\begin{table}[h!]
\centering

\begin{tabular}{m{4.6cm}m{0.7cm}m{0.7cm}}
\toprule
\textbf{Prompt} & \multirow{2}{*}{

\begin{turn}{270}\textbf{\# Tokens}\end{turn}
} & \multirow{2}{*}{

\begin{turn}{270}\textbf{$\mathbf{z}$-score}\end{turn} 
} \\
{Why is it important to measure your heart rate while exercising? What are its practical and advanced applications?
} & & \\
\midrule
\rule{0pt}{2.5ex}\textbf{Baseline} &  &  \\ 
\myred{Heart }\mygreen{rate }\myred{is }\mygreen{the }\myred{number }\mygreen{of }
\myred{times }\mygreen{your }\myred{heart }\mygreen{contracts or }
\myred{beats }\mygreen{per }\myred{minute. }\mygreen{When }\myred{you }
\mygreen{are }\myred{exercising }\mygreen{your }\myred{heart }
\mygreen{be}\myred{ats faster to }\mygreen{deliver more }
\mygreen{blood }\myred{and oxygen }\mygreen{to }\myred{your }
\mygreen{working }\myred{muscles}\mygreen{. }
\mygreen{When you are }\myred{resting your }
\myred{heart }\mygreen{be}\myred{ats slower.}
\mygreen{He}\myred{art rate }\mygreen{can }\myred{be }\mygreen{measured }
\mygreen{using }\myred{a }\mygreen{heart }\myred{rate }\mygreen{monitor }\myred{or }
\mygreen{by manually taking the }\myred{pulse.}
\mygreen{He}\myred{art rate }\mygreen{can }\myred{be }\mygreen{measured }
\mygreen{using }\myred{a }\mygreen{heart }\myred{rate }\mygreen{monitor }\myred{or }
\mygreen{by manually taking the }\myred{pulse.}
& 92 & 5.05 \\ 
\midrule
\rule{0pt}{2.5ex}\textbf{WIS-Perturbation} &  &  \\ 
\myred{Me}\mygreen{as}\myred{uring }\mygreen{heart }\myred{rate }\mygreen{can help }
\mygreen{determine the }\myred{effectiveness }\mygreen{of }
\myred{your work}\mygreen{out, }\myred{as well as }\mygreen{help }
\myred{prevent }\mygreen{inj}\myred{uries}\mygreen{, }\myred{such as }\mygreen{heart }
\mygreen{attacks.}\myred{ It is also used }\mygreen{by }
\mygreen{athlet}\myred{es to determine }\mygreen{the }
\myred{intensity }\mygreen{and effect}\myred{iveness }\mygreen{of }
\myred{their }\mygreen{training.}

\mygreen{There }\myred{are }\mygreen{many }\myred{ways }\mygreen{of }
\myred{measuring }\mygreen{heart }\myred{rate, }\mygreen{including }
\mygreen{a st}\myred{ethos}\mygreen{cope}\myred{, }\mygreen{E}\myred{KG machine }
\mygreen{and }\myred{heart}\mygreen{-}\myred{rate }\mygreen{mon}\myred{itors }\mygreen{(w}\myred{orn }
\mygreen{on }\myred{the chest, w}\mygreen{rists or }\myred{fingers).} 

 & 90 & 5.47 \\ 
\bottomrule
\end{tabular}
\caption{A comparison of texts generated with the baseline watermarking method and WIS-Perturbation. We use $\gamma=0.25$ for both methods. We set $\delta=2.5$ for the baseline, and $\delta=2.75$ for WIS-Perturbation to maintain a similar $z$-score. The text generated with WIS-Perturbation provides a more complete answer.}
\label{tab:example}
\end{table}

\section{Related Work}
\paragraph{Watermarking for LLMs.}
There has been a long history of watermarking machine learning models due to privacy or intellectual property concerns. A series of works \cite{adi2018turning, darvish2019deepsigns} watermark models during training to protect the ownership of the models. 
With the emergence of LLMs, watermarking is used to watermark and identify texts generated by a certain LLM. This is mostly done in the post-training stage for efficiency. These watermarking methods include post-inference watermarking, which modifies the generated texts after the language model produces the whole output \cite{zhang2023remark,yoo2023robust}, and inference-time watermarking, which modifies the logits during each generation step \cite{kirchenbauer2023watermark}. \citet{kirchenbauer2023watermark} proposes a token-level watermarking, which modifies the logits of the next token conditioned on its prefix. \citet{hou2023semstamp} proposes a sentence-level watermarking, which conditions the semantic embedding of the next sentence on its prefix. In this work, we mainly focus on the token-level watermarking \cite{kirchenbauer2023watermark}.

\paragraph{Improving the text quality of watermarked LLMs.}
A few concurrent works are also aware of the downgraded quality of texts generated by watermarked LLMs. \citet{fernandez2023three} evaluates watermarked LLMs on QA tasks, but they are limited to simple QA tasks with short answers.
As a follow-up work of \citet{kirchenbauer2023watermark}, \citet{fu2023watermarking} proposes to expand the greenlist based on the semantic information of tokens, but part of their constructed greenlist is semantically related to the input and shared by all the tokens in the generation, which potentially allows attackers to infer and evade the watermarking mechanism.
\citet{hu2023unbiased} proposes a resampling algorithm to generate texts with watermarking without sacrificing text quality. However, it requires passing the text through the watermarked LLM for reliable detection, which is also inefficient especially considering the size and inference cost of current LLMs. \citet{takezawa2023necessary} proposes an approach that considers the minimal constraints necessary for watermark detection in the generated text, which however increases the time complexity by a factor of the maximum length. Different from these concurrent works, our proposed framework requires only a lightweight model for importance scoring with a small overhead.

\section{Background}
\subsection{Notations}
We consider a generative language model $M$ with a vocabulary $V$. Given a sequence of prefix token $s=[s_1,...,s_{t-1}]$, $s_t=M(s)$ generates the next token with the logits vector $\rvl_t$. Specifically, the logits are normalized to define a probability distribution over $V$, and the probability of token $v_i(1\leq i\leq |V|)$ in $V$ being the $t$-th token in the generated sequence is calculated as $P(s_t \!=\! v_i|s_1,...,s_{t-1})=\frac{\exp(\mathbf{l}_t[i])}{\sum_{1\leq j\leq |V|} \exp(\mathbf{l}_t[j])}$, where $\mathbf{l}_t[i]$ represents the logit corresponding to $v_i$. The final sequence $s=[s_1, ..., s_{T_0}, s_{T_0+1}, ..., s_{T_0 + T}]$ is obtained by autoregressively generating $T$ tokens, following the initial prefix $[s_1, ..., s_{T_0}]$.
    \subsection{Token-level Watermarking and Detection}
In this section, we review the token-level watermarking proposed by \citet{kirchenbauer2023watermark}. 
\paragraph{Generating texts with watermark.}
To generate token $s_t$ at step $t$ with a watermarked LLM, the vocabulary $V$ is randomly partitioned into a greenlist $G$ and a redlist $R$ with a random number generator seeded by the hash value of its prefix $[s_1, ..., s_{t-1}]$. The size of the greenlist is controlled by a hyperparameter $\gamma$, where $|G| = \gamma |V|$. The original logits $\rvl_t$ at $t$ are then modified to $\rvl_t'$ according to the partition of the greenlist and redlist: 
 \begin{align}
 \forall i\in [|V|],\enskip
 \label{eq:logits_ori}
     \rvl_t'[i] = 
     \left\{
     \begin{aligned}
     &\rvl_t[i] + \delta & \text{if $v_i \in G$},\\
     &\rvl_t[i] & \text{otherwise},
     \end{aligned}
     \right.
 \end{align}
 where a hyperparameter $\delta$ is used to control the strength of the watermark.
 The next token at $t$ will then be generated with the modified logits $\rvl'_t$.

\paragraph{Detecting texts generated with watermarked LMs.}
 As described in \eqref{eq:logits_ori}, texts generated with watermarks tend to contain more tokens in the greenlist, which can then be utilized to detect texts generated with watermarked LMs. The detection is done by testing the following null hypothesis:
 \begin{center}
    $H_0$: \textit{Token sequence $s = [s_{T_0+1},...,s_{T_0+T}]$ is generated without watermark,}
\end{center}
with a $z$-test:
\begin{align}
\label{eq:zscore}
z = (|s|_{G} - \gamma T)/\sqrt{T \gamma (1-\gamma)}.
\end{align}
$|s|_{G}$ counts the number of greenlist tokens in the sequence $s$.
We reject $H_0$ and detect $s$ as a text generated by a watermarked LLM if $z$ is larger than a certain threshold which guarantees the false positive rate of the detection.

\section{Proposed Method}
\label{sec:method}
\citet{kirchenbauer2023watermark} watermarks the generated text by increasing the logits of tokens in greenlist, which can potentially harm the quality of generated text if the candidate tokens in greenlist cause worse text quality than the tokens with larger original logits. 
We now describe our proposed method, \textbf{W}atermarking with \textbf{I}mportance \textbf{S}coring (WIS), which utilizes an external importance scoring module to identify and keep important tokens in the generation.

\subsection{Framework Overview}
To start with, we define an importance scoring function $f(s=[s_1, ..., s_{t-1}], s_t): V^* \times V \to [0, 1]$, where $s$ is the input concatenated with output that is already generated at step $t$. $V^*$ denotes the set of sequences with tokens from $V$. We say $s_t$ is predicted as an important token if $f(s, s_t)\geq r_0$ given a pre-defined threshold $r_0$. We then modify the watermarking process described in \eqref{eq:logits} to incorporate this external information about token importance:
\begin{align}
 \label{eq:logits}
     \rvl_t'[i] = \begin{cases}
     \rvl_t[i] + \delta& \text{if } v_i \in G\\
     \rvl_t[i] + \delta & \text{if } v_i = M(s_1, ..., s_{t-1}) \text{and} \\
     &f([s_1, ..., s_{t-1}], v_i) \geq r_0\\
     \rvl_t[i] & \text{otherwise}.
     \end{cases}
\end{align}
where the logits of token $v_i$ is increased by $\delta$ if $v_i$ is in the greenlist, or if $v_i$ should be selected without watermarking and it has an importance score no less than $r_0$.
In the detection, we use the same $z$-test as \citet{kirchenbauer2023watermark}, which is described in \eqref{eq:zscore}, as WIS does not affect the $z$-test against texts generated without watermark.

To implement $f$, we define the importance of a token as the degree to which it contributes to the correctness of output text given an input $s$. We also assume that for a properly pretrained LLM $M$, the original output of $M$ includes tokens that contain important information given $s$. Therefore, we want to identify and keep these tokens that would be generated by $M$ without watermarking. There are several existing methods that aim to estimate the importance of a token given a text context with attention or gradients \cite{madsen2021evaluating,belinkov2019analysis}. However, these methods are computationally expensive if we run the estimation at each generation step. We will introduce two proposed efficient implementations of $f$ in the next sections.

\subsection{Perturbation-Based Importance Scoring}
The perturbation-based importance scoring is based on the idea that if $s_t$ in $s=[s_1, ..., s_t]$ contains important information, the semantic representation of $s$ would be changed significantly if we add a perturbation to $s_t$, such as removing the token. We consider a representation mapping $\phi: V^* \to \mathbb{R}^d$ which maps a text sequence to a $d$-dimensional vector, and we can estimate the importance of $s_t$ given $s$ with
\begin{align}
    &f([s_1, ..., s_{t-1}], s_t) \\
    &= \cos\big(\phi([s_1, ..., s_{t-1}]), \phi([s_1, ..., s_{t}]\big)\nonumber,
\end{align}
where $\cos(\cdot,\cdot)$ computes the cosine similarity between two vectors.
We name the WIS with this perturbation-based importance scoring as WIS-Perturbation.

\subsection{Model-Based Importance Scoring}
\label{sec:model_based}
We can also train a language model to implement $f$, which accepts the sequence $s=[s_1, ..., s_t]$ as the input and outputs the importance score of $s_t$ given $s$. In this case, we need to construct a sequence dataset with importance labeling for each token. 
\paragraph{Constructing a dataset with importance labeling.}
We construct the dataset inspired by the intuition that, if a token $v$ is semantically important given a text $s$, it will stay in the paraphrase of $s$. Therefore, for each text $s^{(i)}$ in a corpus $S = \{s^{(1)}, s^{(2)}, ..., s^{(m)}\}$ with $m$ texts, we paraphrase $s^{(i)}$ for $N$ times with an external paraphrase model. For each token $s_t^{(i)}$ in $s^{(i)}$ with $T^{(i)}$ tokens, we count $n_t^{(i)}$ as the number of appearances of $s_t^{(i)}$ in these $N$ paraphrases. We construct two datasets with importance labels for each token in each text given a corpus $S$:
\begin{align}
    &S_{\text{classification}} \\
    &= \{(s^{(i)}, (\mathbb{I}(n_1^{(i)} > 0), ..., \mathbb{I}(n_{T^{(i)}}^{(i)} > 0)): s^{(i)} \in S \}\nonumber,\\
    &S_{\text{regression}} \nonumber\\
    &= \{(s^{(i)}, ((n_1^{(i)}/N), ..., (n_{T^{(i)}}^{(i)}/N)): s^{(i)} \in S \}\nonumber,
\end{align}
where $S_{\text{classification}}$ with binary labels is constructed for training a classification model and $S_{\text{regression}}$ is constructed for training a regression model. We train the importance scoring model to predict the importance label for each token in the input sequence given its prefix tokens.
We name the WIS with a regression model or classification model as the importance scoring module as WIS-Regression and WIS-Classification, respectively.

\section{Experiments}
In this section, we empirically show that there is a trade-off between generation quality and watermark detection rate. We further show that our proposed WIS methods can produce texts with better quality compared to the baseline token-level watermarking method proposed by \citet{kirchenbauer2023watermark}, with a comparable level of detection rate.

\subsection{Experimental Setup}

\paragraph{Generative Model.}
We employ the LLaMA-2-13B model \cite{touvron2023llama} as the generation model. In all of our following experiments, we set the beam search width to 2 and set the maximum sequence length $T$ to 100.

\paragraph{Datasets.}
We employ two datasets to assess the performance of watermarking methods. The Factual Inconsistency Benchmark (FIB) \cite{tam2022evaluating} contains a modified version of the test set of XSum \cite{Narayan2018DontGM} and CNN \cite{nallapati2016abstractive} datasets. Each sample in the FIB dataset includes a document and a manually revised summary that is consistent with the facts. We use 500 examples from the XSum split in FIB in our experiments.
ELI5 \cite{fan2019eli5} is a long-form question-answering dataset, consisting of a collection of complex questions requiring detailed explanations. We use 500 question-and-answer pairs in ELI5 in the experiments.

\paragraph{Metrics.}
We report the detection rate and ROUGE-1 score \cite{lin2004rouge} on each dataset with different $\gamma$ and $\delta$ settings. Following the settings in \citet{kirchenbauer2023watermark}, we employ a $z$-score threshold of 4.0.

\begin{figure*}[h]
  \centering
  \includegraphics[width=\textwidth]{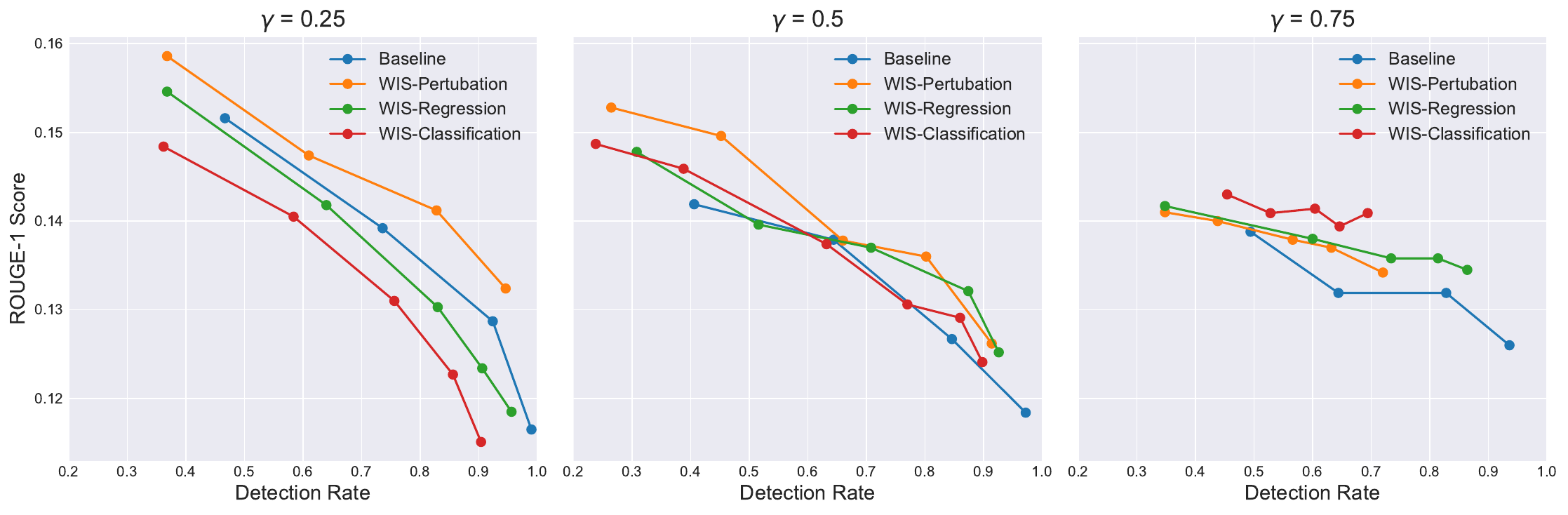}
\caption{Comparison of different scoring functions on the FIB dataset with \(\gamma = \{0.25, 0.5, 0.75\}\). We adjust \(\delta\) to demonstrate the relationship between the ROUGE-1 score and the detection rate. For \(\gamma = 0.25\) and \(\gamma = 0.5\), \(\delta\) is varied within the interval \([1.5, 4]\). For \(\gamma = 0.75\), \(\delta\) is varied within the range of \([2.5, 8]\). We choose the range of $\delta$ to distribute the detection rate between 0.2 and 1.0. We use $r_0$ = 0.02 for WIS-Perturbation and $r_0$=0.9 for WIS-Regression to decide important tokens.} 
\label{fig:FIB}
\end{figure*}

\begin{figure*}[h]
  \centering
  \includegraphics[width=\textwidth]{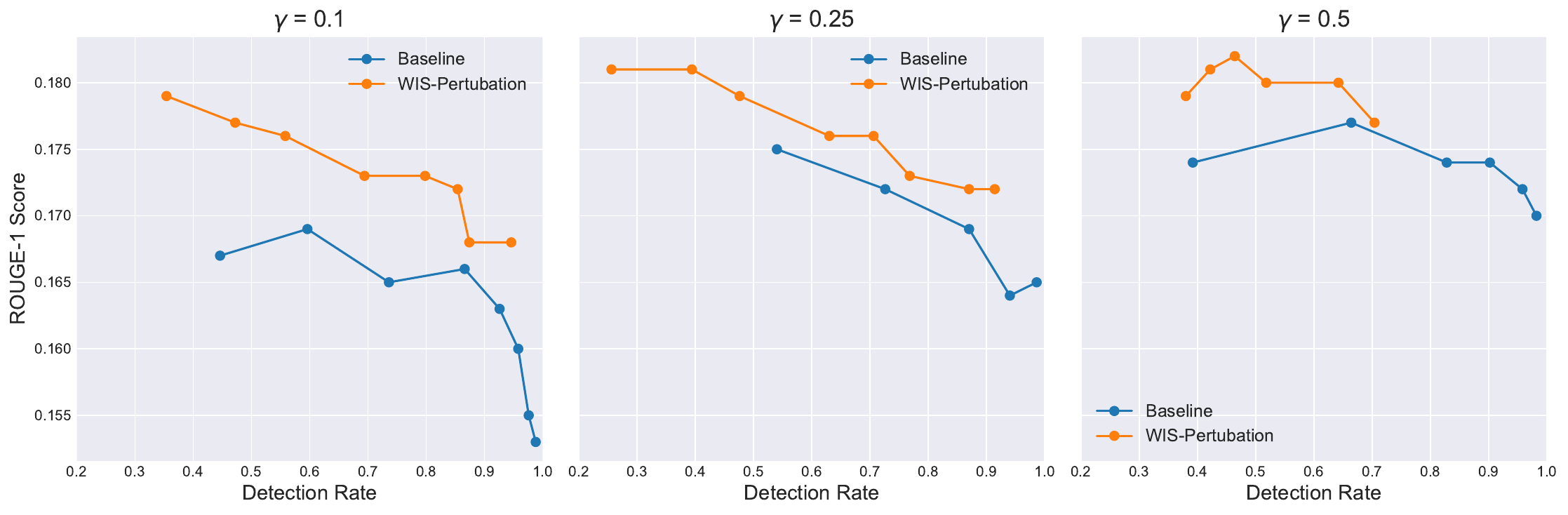}
  \caption{Comparison between perturbation-based importance scoring and the baseline on the ELI5 dataset with \(\gamma = \{0.1, 0.25, 0.5\}\). For the baseline method, \(\delta\) is varied within the interval [1, 2.75], and for WIS-Perturbation, \(\delta\) is varied within the range of [1.25, 3.75]. We choose the range of $\delta$ to distribute the detection rate between 0.2 and 1.0. We use $r_0=0.02$ for WIS-Perturbation.}
  \label{fig:ELI5}
\end{figure*}

\paragraph{Importance scoring function implementation.}

Following Section \ref{sec:method}, we implement three importance scoring functions, including WIS-Perturbation, WIS-Regression and WIS-Classification.
For WIS-Perturbation, we leverage BERTScore \cite{zhang2019bertscore} to compute the cosine similarity between embeddings with a pretrained BERT-base model as the representational mapping $\phi$. For WIS-Regression and WIS-Classification, we construct the datasets as described in Section \ref{sec:model_based} using 180,000 examples from the XSum training dataset. For WIS-Regression, we fine-tune a BERT-base model \cite{devlin2018bert} with an additional fully connected layer and a sigmoid activation function. For WIS-Classification, we fine-tune another BERT-base model, augmented with an extra fully connected layer for binary classification.
To enhance the efficiency during inference, we incorporated a sliding window mechanism on the input sequence $s$ for all three importance scoring functions:

\begin{align}
    &f(s,s_t)=f([s_{t-w}, ..., s_t], s_t),
\end{align}
where the window size $w$ is set to 16 for FIB and 10 for ELI5.

\subsection{Trade-off between Detection Rate and Generation Quality}
\label{sec:trade-off}

We report the results from FIB dataset and ELI5 dataset in \Cref{fig:FIB} and \Cref{fig:ELI5}, respectively. We group results from different $\gamma$ and $\delta$ combinations with different $\gamma$, as the calculation for $z$-score is the same with the same $\gamma$.

We can observe a clear trend of trade-off between generation quality (reflected by ROUGE-1 scores) and the detection rate on both datasets. 
With the increment of $\gamma$, the reduction in ROUGE-1 score brought by the increase in detection rate decreases, indicative of a lesser impact caused by watermarking on the generated text's quality. This effect is most pronounced at higher $\gamma$ levels, where the green list is increasingly likely to include tokens with the highest original logits. 

\subsection{Improved Generation Quality with Importance Scoring}
Comparing across the FIB and ELI5 datasets, all of our three methods, including WIS-Perturbation, WIS-Regression and WIS-Classification, have demonstrated superior performance over the baseline. As depicted in \Cref{fig:FIB}, WIS-Perturbation outstrips the baseline in achieving higher ROUGE-1 scores at comparable detection rates, showcasing its efficacy notably at $\gamma=0.25$. With an elevation in $\gamma$, WIS-Regression and WIS-Classification begin to dominate. For the ELI5 dataset, \Cref{fig:ELI5} shows that WIS-Perturbation consistently provides better text quality than the baseline at matching detection rates. Nonetheless, the maximum detection rate achievable by WIS-BERTScore is decreased as $\gamma$ is incremented, with a maximum detection rate of 70.4\% at $\gamma=0.5$. This is due to the fact that with the increment of $\gamma$, the number of necessary green list tokens $|s|_G$ for the text to be detected increases. However, our method retains tokens from the redlist to improve the text quality, which limits the detection rate under a large $\gamma$

\section{Conclusion}
This paper presents a new framework, Watermarking with Importance Scoring (WIS), designed to improve the trade-off between text quality and detection efficacy in LLM watermarking. WIS introduces an external importance scoring function, which aims to preserve tokens critical for the accuracy of the generated text at inference time. The experimental results on several datasets demonstrate that, with both perturbation-based and model-based importance scoring methods, WIS significantly enhances text quality while maintaining a comparable rate of watermark detection. 

\bibliography{papers,custom,tacl}

\appendix

\end{document}